\def\year{2021}\relax
\documentclass[letterpaper]{article} % DO NOT CHANGE THIS
\usepackage{aaai21}  % DO NOT CHANGE THIS
\usepackage{times}  % DO NOT CHANGE THIS
\usepackage{helvet} % DO NOT CHANGE THIS
\usepackage{courier}  % DO NOT CHANGE THIS
\usepackage[hyphens]{url}  % DO NOT CHANGE THIS
\usepackage{graphicx} % DO NOT CHANGE THIS
\usepackage{natbib}  % DO NOT CHANGE THIS AND DO NOT ADD ANY OPTIONS TO IT
\usepackage{caption} % DO NOT CHANGE THIS AND DO NOT ADD ANY OPTIONS TO IT
\usepackage[hyphens]{url}
\usepackage{hyperref}
\usepackage[hyphenbreaks]{breakurl}

\usepackage{booktabs} % For formal tables
\usepackage{amsmath}
\usepackage{physics}
\usepackage{multirow}
\usepackage[switch]{lineno}  %
\usepackage[dvipsnames]{xcolor}

\title{Unsupervised Domain Adaptation for Hate Speech Detection\\ Using a Data Augmentation Approach}
\author{Sheikh Muhammad Sarwar,\thanks{Work done while the author was an intern at Amazon.com}\textsuperscript{\rm 1}
 Vanessa Murdock,\textsuperscript{\rm 2}\\
 } 
\affiliations{%
 \textsuperscript{\rm 1} CIIR, College of Information and Computer Sciences\\University of Massachusetts Amherst \\
 \textsuperscript{\rm 2} Amazon.com\\
 smsarwar@cs.umass.edu, vmurdock@amazon.com
 }
\begin{document}
%\linenumbers  %

\maketitle

\begin{abstract}
 %Online harassment frequently uses hate speech and has been increasing in recent years. Addressing the problem requires a combination of content moderation by humans and automatic hate speech detection.  

Online harassment in the form of hate speech has been on the rise in recent years. Addressing the issue requires a combination of content moderation by people, aided by automatic detection methods. As content moderation is itself harmful to the people doing it, we desire to reduce the burden by improving the automatic detection of hate speech.  Hate speech presents a challenge as it is directed at different target groups using a completely different vocabulary.  Further the authors of the hate speech are incentivized to disguise their behavior to avoid being removed from a platform.  This makes it difficult to develop a comprehensive data set for training and evaluating hate speech detection models because the examples that represent one hate speech domain do not typically represent others, even within the same language or culture.  We propose an unsupervised domain adaptation approach to augment labeled data for hate speech detection. We evaluate the approach with three different models (character CNNs, BiLSTMs and BERT) on three different collections.  We show our approach improves Area under the Precision/Recall curve by as much as 42\% and recall by as much as 278\%, with no loss (and in some cases a significant gain) in precision. 
\end{abstract}


%\keywords{Hate Speech Detection, Domain Adaptation}

% \input{original_submission/Introduction.tex}
% \input{original_submission/literature.tex}
% \input{original_submission/problem_formulation.tex}
%  %\input{approach.tex}
%  \input{original_submission/experiments.tex}
%  \input{original_submission/conclusion.tex}

\section{Introduction}
Online harassment in the form of hate speech has been on the rise in recent years. A recent paper~\cite{adl2020} from the Anti-Defamation League\footnote{\url{https://www.adl.org} visited May 2021} reports that nearly half (44\%) of Americans report having experienced some type of online harassment, up from 41\% in 2017. Of those, 35\% report having been harassed as a result of their sexual orientation, religion, race or ethnicity, gender identity, or disability. 

The problem is exacerbated by machine learned systems that are trained using labeled data from online forums.  With inadequate hate speech filtering, these systems themselves become vectors of hate.  For example, YouTube~\cite{yt2020} has been found to promote hate speech via its recommended videos simply by learning from user interactions. In 2016 Microsoft released a conversational agent ``Tay" that learned from user interactions on Twitter, but had to take it down a short time later because it was generating racist content~\cite{tay2016}. 

To filter hate speech, a machine learned system will need large amounts of training data with adequate coverage of the vocabulary.
It is difficult to create a high-coverage vocabulary of offensive terms or phrases that occur in hate speech mentions because of regional and linguistic variants even within the same language, compounded by variety in the targets of hate speech. The terms directed at one target often have little or no overlap with terms directed at a different target. Furthermore, hate speech often does not contain any terms that are offensive in and of themselves. Rather it is contextually hateful, referring to offensive stereotypes, or alluding to or inciting violence against a target group.

Recent approaches to hate speech detection are based on supervised neural representation learning~\cite{macavaney2019hate, glavas-etal-2020-xhate, pamungkas-patti-2019-cross, badjatiya_www19, agrawal18, Arango:2019, Waseem2018multitask}. These approaches require a large number of hate speech instances to achieve high recall in the hate speech class. \citet{Arango:2019} found that the performance of neural models trained using data from \citet{waseem:2016} drops significantly when tested on data from \citet{semeval}, which is from a different domain. The failure of the models to generalize to a target domain is due to user bias in the source domain data, where a small number of users generate the majority of hateful examples. Furthermore, since hate speech occupies a tiny proportion of data from a domain, test collections are often constructed by searching with a set of seed terms from a hate speech lexicon.  This results in a data set with a domain-limited vocabulary which itself may have the shortcomings noted above. For example, a source data set seeded by anti-Muslim terms may be inadequate for detecting anti-Woman content in target domain data.

One way to address the domain mismatch is to gather labeled data from the target domain. Since it is sensitive and costly to obtain annotations for hate speech ~\cite{schmidt, waseem:2016, malmasi2018challenges, mathur2018detecting}, it is desirable to utilize unlabeled data from the target domain to build a robust classifier. Thus, Unsupervised Domain Adaptation (UDA) -- i.e., the problem of building a robust target domain classifier with labeled data from the source domain and unlabeled data from the target domain -- is a realistic and important problem in the context of hate speech detection. 

We contribute a method that automatically generates a domain-adapted corpus to bridge the gap between source domain and target domain for  hate speech detection.  Although there are cross-domain studies for hate speech detection, to the best of our knowledge, this is the first study of UDA for hate speech detection.

 We identify hate speech sentences where the hate speech content terms can be distinguished from their surrounding sentence context.  For example\footnote{In this paper we intentionally use non-hate examples to limit the level of offensiveness in the paper itself.  In this example ``Honda CRVs" (or by proxy, their owners) are not considered an at-risk or protected group.} in the sentence ``The problem with Honda CRVs is that they are boring", the content consists of the subject ``Honda CRVs" and the negative descriptor ``boring".  The surrounding sentence context is ``The problem with ... is that they are ...".

While all hate speech does not have this structure, leveraging examples that do provides a convenient template for domain adaptation. We can automatically identify the template in generic sentences with negative sentiment. and slot in hate speech content to convert it to hate speech in a new domain.  Note that the process does not have to be perfect because this type of training data can be generated in large quantities.  

To create a domain-adapted corpus, we train a sequential tagger on the labeled data in the source domain so that the tagger is able to identify hate speech content terms, and surrounding sentence context templates. We apply the tagger to unlabeled data in the target domain to derive a lexicon of hate terms in the target domain. We also apply it to a large corpus of generic sentences with negative sentiment. This yields a large data set of sentence contexts that will serve as hate speech templates. In this work we use a collection of Twitter posts labeled with negative sentiment based on emojis~\cite{go2009twitter}. As the posts are labeled using emojis this collection can be extended without any supervision meaning we can generate hate speech templates in abundance. 

To adapt the generic hate speech templates to the target domain, we rank them according to their textual similarity to the target domain sentences, and select the top $k$ for augmentation. This reduces noise in the domain-adapted data and increases the topical similarity between the generic templates and the target domain. Finally, we impute terms from the derived hate speech lexicon from the target domain into the generic templates.  The result is a large corpus of negative sentences with hate speech content from the target domain.  The details of this process are explained in Section~\ref{sec:approach}.    

We evaluate the approach on three different models:  character CNNs, BiLSTMs and BERT; and on three different collections: Semeval~\cite{semeval}, Gibert~\cite{gibert} and Hasoc~\cite{hasoc2019overview}.  We show that using a  domain-adaptation approach to augment the training data with synthetic data in the language of the new domain, we are able to improve hate speech detection across the board by as much as 42\% AUCPR, and 278\% recall with the Hasoc data, and 8\% improvement in AUCPR and 27\% recall with Semeval data, and 14\% improvement in AUCPR with Gibert data.

% \sheikh{The summary of our contributions are:
% i) We address the UDA problem for domain adaptation for the first time in the literature. 
% ii) We create a UDA setting from the available datasets. iii) We propose a technique to generate domain adaptive training data by leveraging the vocabulary of the target domain unlabeled data. We use weakly labeled negative emotion sentences. We can potentially generate infinite amount of hate speech data. iv) we show that data augmentation based UDA technique performs better on two out of three data sets compared to a strong baseline.
% }

The rest of the paper is structured as follows.  Section~\ref{literature} surveys the current literature in bias in hate speech detection, and domain adaptation.  Section~\ref{sec:approach} presents our approach to domain adaptation for hate speech.  Section~\ref{data} discusses the existing collections for research in hate speech detection.  Section~\ref{experiments} presents the experimental set up.  Section~\ref{discussion} discusses the results.%, 
\section{Related Work}\label{literature}

%\sheikh{Reread for language because we on-the-fly edited some things. }
%\vanessa{I think we can do a better job of related work on hate speech detection.  We should add the following related work: Viviana Patti has a number of papers recently about hate speech detection.  I just queried google scholar and found a bunch that we could cite/discuss }\\
%\vanessa{R3 mentions these papers
%Waseem, Zeerak, James Thorne, and Joachim Bingel. Bridging the gaps: Multi task learning for domain transfer of hate speech detection. Online harassment. Springer, Cham, 2018. 29-55.\\
% Karan, Mladen, and Jan Šnajder. Cross-domain detection of abusive language online. Proceedings of the 2nd workshop on abusive language online (ALW2). 2018.
%}

Hate speech detection is a relatively recent research area, gaining interest from 2015.  Prior work focuses on offensive content detection (for example~\citet{yingchen}) and opinion mining~\cite{bingliu}. One of the early papers specifically focused on hate speech~\cite{hirshberg} defines hate speech as containing hateful content directed at a protected group, which is similar to the hate speech template employed in this paper.  While there is a growing body of literature on approaches to hate speech detection (c.f.~\citet{macavaney2019hate} and ~\citet{schmidt}), we discuss the literature on data for hate speech detection, and domain adaptation, as the focus of this paper is data augmentation for hate speech, assuming there is only unlabeled data from the target domain. 

The robust labeling approach proposed by \citet{founta2018large} focuses on fine-grained abusive behavior detection, treating it as a multi-class classification problem. They applied several techniques for obtaining robust labels from annotators, but did not apply any automatic approach specifically for hate speech detection. They used random boosted sampling to obtain a large collection of samples for human annotation. We propose an automatic method to generate labeled samples from a large collection of negative emotion sentences~\cite{go2009twitter}, as we wish to reduce the reliance on human annotation.
%\subsection{Bias in Hate Speech Datasets}
%\vanessa{we need to add a preamble honoring waseem and citing both his papers}

One of the most studied public data sets labeled for hate speech was introduced in a pair of papers by \citet{waseem:2016} and \citet{cybertwitter}.  This work provided a test bed and a methodology for studying hate speech.  Because it is one of the first data sets, it is also one of the most studied, and subsequent work elucidated bias and other issues common in hate speech detection using this collection.

In general most hate speech datasets are biased because of the sampling procedure. \citet{wiegand-etal-2019-detection} demonstrated that a common method for sampling data for hate speech detection (focused sampling) results in datasets biased toward author and topic. %Weigand et al. took one of the most meticulously documented datasets constructed by \citet{waseem:2016}, and showed that focused sampling resulted in author as well as topic bias. %The authors provided a summary on both types of biases on different datasets.  
%Wiegand et al. also mention random boosted sampling but we don't discuss it so I omitted it.
Topic bias results in a domain specific dataset. The dataset provided by \citet{waseem:2016} contains tweets mostly about women in sports with a focus on their competence as football commentators. \citet{wiegand-etal-2019-detection} showed that the data contains domain-specific keywords such as \emph{announcer}, \emph{commentator}, \emph{football}, \emph{sports}, occurring frequently in the data as a whole, and specifically in the abusive tweets. 

Apart from topic bias, Weigand et al. found that two authors \textbf{Male tears \#4648} and \textbf{Yes, They're Sexist} generated more than 70\% of the sexist tweets, while a single author \textbf{VileIslam} generated 90\% of the racist tweets. Overall the authors reported that a focused sampling strategy made the Waseem data domain- and user-style specific. The authors suggested that it is imperative to perform cross-domain classification to analyze the predictive power of a model constructed from any hate speech data.  

To analyze the predictive power of the data set by \citet{waseem:2016},  \citet{Arango:2019} performed a cross-dataset analysis having the Waseem data as the source and empirically demonstrated the effect of biased training data. They trained a BiLSTM model adopted from \citet{agrawal18} on the Waseem data, tested the model on the Semeval dataset~\cite{semeval}, and discovered an extreme drop in performance. The bias in the Waseem data arises because only 1,590 users write all the tweets in the collection. Moreover, a fine-grained analysis discovered that 491 users generated all the sexist tweets, while only 8 users generated all the racist tweets. Even worse, a single user generated 40\% of all the sexist tweets, and another individual user generated 90\% of all racist tweets. These findings were consistent with the findings of \citet{wiegand-etal-2019-detection}. \citet{waseem:2016} also mentioned that the inter-annotator agreement is $\kappa$ = 0.84 and all disagreements occur in  annotations  of  sexism. This suggests that the racist examples were very straightforward and therefore less valuable for training a model. %This analysis provided by \citet{Arango:2019} showed the nature of bias that exists in \cite{waseem:2016}'s dataset.   

%In this paper, we refer the SemEval 2019 Task 5 about the detection of hate speech against immigrants and women in Spanish and English messages extracted from Twitter as the Semeval-dataset \cite{semeval}. 
\citet{Arango:2019} showed that cross-dataset performance can be improved by removing bias from the training data, and adding data from another source (in this case the hate speech data provided by \citet{davidson}). However, it is not clear whether the performance gain achieved by Arango et al. is caused by de-biasing or augmenting the data. Moreover, this cross-dataset experimentation was not complete. Typically domain-adaptation studies evaluate a model trained from a source domain across more than one target domain.  

\subsection{Domain Adaptation (DA)} 
Machine learning models assume that the same underlying distribution generates the source and target domain data. However, this assumption is not true for all applications \cite{Daume:2006}. %For example, we have a spam detector trained on the emails of a set of users (source domain) and want to adapt id for a new user (target domain). However, it might happen that the emails for the new users possess very different features compared to the emails in the training data. 
In fact, it has been shown that the source and the target domains come from different distributions for many tasks %This situation happens in a lot of tasks including named entity recognition 
including named entity recognition \cite{linlu2018, tian2016}, sentiment classification \cite{blitzer2007}, and information retrieval \cite{cohen2018cross,tran2019domain}. 

Domain adaptation techniques can be classified into \emph{supervised} and \emph{unsupervised} \cite{daume2006}. In terms of supervised approaches, \citet{rizoiu2019transfer} considered accessing 90\% of the source and target domain data to predict 10\% of the target domain data, which might not always be practical. %Karan et al.\cite{karan2018} adopted the frustratingly easy domain adaptation approach for hate speech detection which is also supervised . 
\citet{sharifirad2018} applied a text generation approach based on a knowledge-base to generate more source domain data. For example, their approach replaced a source domain keyword ``girl" with the word ``woman" using the ``Is-A" relationship from ConceptNet\footnote{\url{https://conceptnet.io/} visited May 2021}. Their generation approach is lexical rather than topical.  %which can be achieved by word embedding based classification models. 
Moreover, their approach does not leverage unlabeled data from the target domain. %Therefore, to the best of our knowledge a study on a challenging unsupervised domain adaptation setting is not yet performed yet. 
%These supervised domain adaptive hate speech detection approaches do not shed light on more recent model architectures that are available today. We provide a comparison of character-based \cite{kim2014convolutional}, word-piece based \cite{bert2018} and word-based models \cite{agrawal18} for domain adaptation in Section~\cite{}. 

Unsupervised Domain Adaptation (UDA) considers labeled data in a source domain and unlabeled data in a target domain, which more closely reflects ``real world'' applications \cite{ruder2019neural}. UDA techniques have been applied to many text classification tasks, but most relevant to the current work, sentiment analysis tasks \cite{xue2020improving, hu2019domain, he-EtAl:2018, chen-cardie-2018-multinomial, zhang2019interactive, acan:2019}. All these approaches focus on extracting domain-independent features from both source and target domain data, using labels from source domain data to learn a sentiment classifier on the features. 

\citet{he-EtAl:2018} devised a semi-supervised approach to use target domain data to train a sentiment classifier.
\citet{hu2019domain} proposed to distill domain-independent features by adding a domain-dependent task that strips out domain-dependent features. \citet{acan:2019} proposed a category alignment approach to avoid ambiguous target domain features near the decision boundary of the sentiment classifier and achieved state-of-the-art results for cross-domain sentiment classification. We adapted this approach to hate speech detection, and show the performance in Table~\ref{tab:acan}.  %\sheikh{We adopt the model from \citet{acan:2019} for hate speech detection and use it as a baseline. This does not connet with the next paragraph may be?}

While all these approaches focus on learning domain-invariant representations and calibrating classifier decision boundaries to perform better classification in the target domain for sentiment classification, there has been no study of their applicability to unsupervised cross-domain hate speech detection. There are a few studies that report cross-domain performance of different abusive content detection models, but they do not provide any direction to make these models adaptable using unlabeled data from the target domain~\cite{glavas-etal-2020-xhate, pamungkas-patti-2019-cross, karan2018}. 

\citet{karan2018} discuss the difficulty of UDA for hate speech detection, in particularly that it is necessary to have some in-domain training data. They did not address the UDA problem and used the Frustratingly Simple Domain Adaptation (FEDA) technique from \citet{daume-iii-2007-frustratingly} with labeled data from the target domain.  

\citet{Waseem2018multitask} proposed a multi-task learning approach to integrate different datasets into a single training process to construct a generalized hate speech detection model. As this approach also uses labeled samples from all the datasets in both training and evaluation, it does not tackle the UDA problem, where no labeled data from the target domain exists. We create a UDA setting and proposed a data augmentation based UDA approach for hate speech detection that applies semi-supervision on a sentiment analysis data set and does not require learning of domain-invariant features.

\section{UDA for Hate Speech Detection}
\label{sec:approach}

As mentioned above, hate speech detection has a bias problem where a classifier might learn the hate speech vocabulary and usage patterns of a very small number of people, and be unable to generalize to hate speech in a new domain, directed at other groups.  One solution is to limit the contribution of any given individual, as in \citet{Arango:2019}. We found that increasing the amount of training data is also effective even without limiting the contribution of an individual (further discussed in Section~\ref{prelim}). However, neither solution solves the problem of adapting to a new domain.  We propose an UDA approach that both augments the training data, and adapts to the target vocabulary.

\emph{Problem Setting} We have a source domain hate speech dataset $D^s$ 
with labeled examples, and unlabeled data $D_u^t$ from the target domain. The task is to train a hate speech detection model using $D^s$ and $D_u^t$.   We evaluate it on the labeled data from the target domain $D^t_l$.

We augment the source domain dataset, $D^s$, with domain-adapted hate speech in the target domain. 
We describe the process in detail below, and an example sentence transformation for each step is shown in Table~\ref{steps}.

\begin{table*}
\begin{center}
\begin{tabular}{|c|l|l|}\hline
\bf{Symbol} & \bf{Explanation} & \bf{Example}\\\hline
$D^s_{hate}$ & Hate Speech in the source domain & The problem with Honda CRV's is that they are boring \\
$H^s$ & Source domain (external) hate lexicon of OTG tokens & honda, crv, boring \\
& Context carrier & The problem with ... is that they are ... \\
$\widetilde{D}^s$ & Templatized sentence used to train an OTG tagger & The problem with REP is that they are REP \\\hline
$D^t_u$ & Unlabeled data from the target domain & Bananas are very yucky!\\
$H^t$ & Target domain lexicon of OTG tokens derived from tagging & bananas, yucky\\
$\widetilde{D}^t_u$& Templatized target domain sentence (for similarity scoring) & REP are very REP!\\\hline 

$D^{weak}$& Negative emotion sentence & I hate Sundays -- they are so dull \\
$\widetilde{D}^{weak}$ & Negative emotion sentence after tagging and templatizing & I hate REP -- they are so REP\\
& Negative emotion sentence, domain adapted & I hate bananas -- they are so yucky\\\hline
\end{tabular}
\caption{Example sentences from each stage of the domain adaptation.  The hate speech lexicon used to derive token-level labels in the source data is from an external source, whereas the hate lexicon for the target domain is the result of applying the tagger to the unlabeled target domain data.  The negative emotion sentences are generic and are not related to either the source or the target domains.  They are adapted to the new domain first by selecting the sentences that are most topically similar to the target domain, and then imputing target domain hate speech tokens into the sentences.}\label{steps}
\end{center}
\end{table*}

\subsection{Learning a Tagger From the Source Domain Data }
\label{sec:otg_tagger}
We define \emph{context carriers}, which contain useful patterns from which a variety of hate speech can be generated. For example in the sentence ``The problem with Honda CRVs is that they are boring" the \emph{context carrier} is ``The problem with ... is that they are ...". We also define \emph{Offensive or Target Group} (OTG) tokens as combination of offensive keywords and keywords indicating a specific race, gender, religion, etc. that are the target of the offense. These are the hate speech \emph{content} of a sentence. We learn an OTG token tagger, $T_{OTG}$, from the source data $D^s$, that outputs hate speech content and context carriers from a sentence input. 

The data $D^s$ is labeled for sentences rather than tokens, but almost all the hate speech datasets are retrieved from social media or blog search systems with queries from a hate speech lexicon.  In this paper we used the lexicon from hatebase.org\footnote{\url{https://hatebase.org/} visited May 2021} as the hate speech lexicon, $H$.   Entries in $H$ are unigrams (such as ``criminal'') and phrases that mention offensive terms and a target group.  
We tokenize the phrases and consolidate them with the unigrams to create a lexicon of OTG tokens, $H^s$.

To create training data for $T_{OTG}$, we select examples from $D^s$ that have been labeled as hate speech at the sentence level, $D^s_{hate}$. 
We iterate over the tokens in $D^s_{hate}$, and label tokens  as ``OTG'' that have a match in the hate lexicon $H^s$. Other tokens are labeled as ``O''. We did not use non-hate examples from $D^s$ for training the model even if OTG tokens appear in that part of the data, because the appearance of OTG tokens in a neutral sentence is not necessarily indicative of offensiveness or hate.  
For example, a sentence might mention the International Criminal Court, matching the hate term ``criminal'' and not be in any way offensive or hateful.  

Once we label the sequence tagging data set from the source hate speech data set, we learn the sequence tagger, $T_{OTG}$. We used both character and word level representations in the model. The character-level representation captures terms that have been encoded\footnote{Encoding substitutes numbers and special characters for letters in words to evade lexical pattern matching.} to avoid automatic detection.  

\begin{figure}
    \centering
    \includegraphics[width=0.9\linewidth]{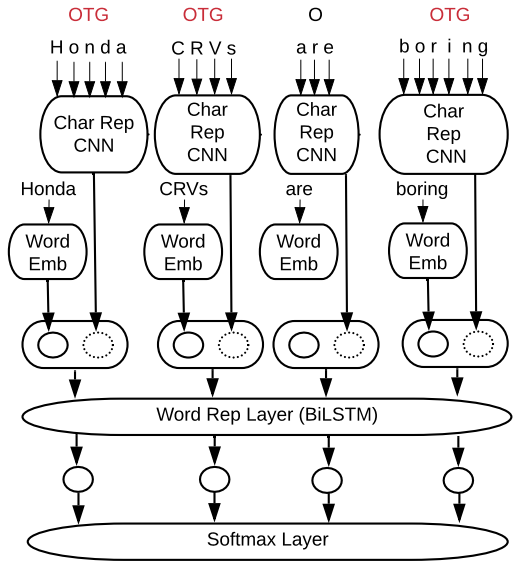}
    \caption{The Offensive or Target Group (OTG) tagging model. The model makes use of character-level and word-level information.  In this example ``Honda'' and ``CRVs'' are the Target, ``boring'' is offensive, and ``are'' is neutral.  Tokens are labeled ``OTG'' and ``O'' accordingly.}
    \label{sheikhFig1.png}
\end{figure}

The tagger $T$ encodes character vectors using convolutions, and then max pooling obtains the character-based representation of a word. The word embedding representation is concatenated with it. A Bidirectional Long Short Term Memory (BiLSTM) layer is applied on top of the concatenated representations to obtain a contextual word representation. Finally, a Softmax layer is applied on the word representation to obtain a probability distribution over the label set. 
An example is shown in Figure \ref{sheikhFig1.png}. The tokens ``honda", ``CRVs" and ``boring" are tagged as OTG tokens.

To create a weakly-labeled data set in the target domain, 
we apply the tagger $T$ to the (unlabeled) target domain dataset, $D^t_u$. This produces two outputs: a new hate speech lexicon comprised of OTG tokens in the target domain, $H^t$, and the set of target domain context carriers $\widetilde{D}^t_u$.  We replaced the OTG tokens with the token ``REP'' to templatize the sentences.  Note that the context carrier now represents the topic of the sentence, minus the hate terms.

We also apply the tagger to the noisy negative emotion data set $D^{weak}$ to obtain the negative emotion context carriers $\widetilde{D}^{weak}$, which we also templatize with the token ``REP". We discard the tokens tagged as ``OTG" in the negative emotion data because they are more likely to be generic nouns and adjectives.

\subsection{Adaptation of Weakly Labeled Data to the Target Domain}
The above process yields a large weakly-labeled corpus of synthetic hate speech $\widetilde{D}^{weak}$ candidate sentences, which are unrelated topically to either the source or target domains. We adapt this corpus to the target domain as follows.
We represent the sentences in both $\widetilde{D}^{weak}$ and $\widetilde{D}^t_u$ as tf-idf vectors. For each sentence in $\widetilde{D}^{weak}$ we compute the cosine similarity to each sentence in $\widetilde{D}^t_u$.  This produces a vector of similarity scores for each sentence in $\widetilde{D}^{weak}$, which we sum to produce a single score which represents the topical similarity of the sentence to the target domain. Note that this similarity is computed in the absence of OTG tokens.

We select the top 10,000 sentences according to the similarity score that contain at least two ``REP'' tokens. We replace the ``REP'' tokens with tokens from the target domain hate lexicon $H^t$, uniformly and at random.  We label these sentences as hate speech.  Random sampling is a reasonable strategy here because it reduces bias towards any specific OTG term.  Note that although the tagger was trained entirely on hate speech sentences, there is no guarantee a whether a specific term in $H^t$ is offensive or target group indicative. This work is focused towards creating robust out-of-domain hate speech detectors without any additional labeled data.  

We also select the top 10,000 sentences that contain no more than one ``REP'' token, and replace all ``REP'' tokens with tokens randomly sampled from $H^t$. We label these sentences as non-hate speech to allow the learner to distinguish between hate speech (directed at a target) and speech which is merely offensive. The final data set is comprised of the labeled source dataset $D^s$, and the domain adapted training sentences, containing both hate and non-hate examples.     

\begin{table*}
\begin{center}
\begin{tabular}{c|c|c|c}\toprule
\bf {Dataset name} & \bf{Number of} & \bf{Hate} &  \bf{Source of Data}\\
& \bf{examples} & \bf{Speech} &  \\\hline\hline
\textbf{WA}~\cite{waseem:2016} & 14949 & 4839 &  Tweets \\\hline
\textbf{DBW}~\cite{davidson} & 24783 & 4993 & Tweets \\\hline
\textbf{SE}~\cite{semeval} & 9000 & 3783 & Tweets \\\hline
\textbf{GI}~\cite{gibert} & 10944& 1196 & Forum posts \\\hline
\textbf{HA}~\cite{hasoc2019overview} & 5852 & 1143 & Facebook posts and tweets \\ \hline
\textbf{AR}~\cite{Arango:2019} & 7006 & 2920 & Unbiased WA and DBW hate speech
\\\bottomrule
\end{tabular}
\caption{Description of the hate speech datasets}\label{tab:data_stat}
\end{center}
\end{table*}

\section{Hate Speech Datasets}\label{data}
 
We consider the data sets provided by \citet{waseem:2016} and \citet{davidson} as source data following \citet{Arango:2019}. We included two more data sets as target data sets provided by \citet{gibert} and \citet{hasoc2019overview} along with the only data set provided by \citet{semeval} that \citet{Arango:2019} used in their experiments. Note that we create a UDA setting from all these data sets, which we describe in Section \ref{sec:uda_setting} and this section only provides a summary of the original data sets.  Table \ref{tab:data_stat} provides the collection statistics for the data sets.    

\subsection{Source Domain Data}

\textbf{WA:} \citet{waseem:2016} 
collected 136,052 tweets, from two months of Twitter\footnote{www.twitter.com visited May 2021} data, focusing on entities likely to engender hate speech. They annotated 16,914 of the tweets. A tweet is annotated as hate speech if it uses a sexist or racial slur, or attacks a group of people on the basis of their religion, gender, ethnicity or sexuality, or if it defends xenophobia or sexism. Their  specific approach to collection and annotation ensured that non-hate speech in this corpus contains offensive terms. These offensive examples that are not hate speech present a challenge to hate speech detection because it is difficult for a classifier to distinguish the hateful tweets from those that are merely offensive. %For example, consider the following tweet mentioned in their paper:

\noindent
\textbf{DBW:} \citet{davidson} queried twitter using a hate speech lexicon from \url{hatebase.org} and retrieved 85.4 million tweets written by 33,458 users. From this large collection they randomly selected 25k tweets and crowd-sourced the annotations as one of three categories: hate speech, offensive but not hate speech, or neither offensive nor hate speech. They defined hate speech as a language used to express hatred towards a targeted group or intended to be derogatory, to humiliate, or to insult the members of the group.
 Although the tweets were retrieved using offensive keywords, only 5\% of the randomly sampled tweets were coded as hate speech, while a majority of them were identified as offensive.         

\noindent
\textbf{AR:} \citet{Arango:2019} de-biased WA \cite{waseem:2016} and added hate speech tweets from DBW \cite{davidson} to create a combined dataset that outperformed models trained on the biased WA by a large margin using SE \cite{semeval} as the test set. Because of this improvement over the previous data sets, and its focus on domain bias, AR is our baseline dataset, and the base upon which we augment the data. %\looseness-1    

\subsection{Target Domain Data}

\textbf{SE:} \citet {semeval} released this dataset for the ``Multilingual detection of hate speech against immigrants and women in Twitter'' task at SemEval. The task organizers defined hate speech as ``any communication that disparages a person or a group on the basis of some characteristic such as race, color, ethnicity, gender, sexual orientation, nationality, religion, or other characteristics.''  Tweets were collected using multiple strategies including monitoring the accounts of people known to use hate speech, as well as sampling tweets containing terms from a lexicon of offensive keywords. The dataset is multi-lingual (Spanish and English). The English training set consists of 10,000 tweets among which roughly 40\% represent hate speech. %Frequent keywords from this collection are:migrant,refugee,\#buildthatwall, bitch, hoe, women.

\noindent
\textbf{GI:} \citet{gibert} sampled sentences published between 2002 and 2017 collected from Stormfront, a white supremacist forum. It contains 10,568 sentences classified into hate speech and non-hate speech. The annotators define hate speech as ``a deliberate attack directed towards a specific group of people motivated by aspects of the group's identity."

\noindent
\textbf{HA:} \citet{hasoc2019overview} created a labeled collection of posts from Twitter and Facebook in Indo-European Languages: German, English, and Hindi. The organizers created evaluation benchmarks for three sub-tasks, and we use labeled data for the binary classification task that requires a model to classify a post as hate speech or non-offensive. We use the training dataset for English in our evaluation. After manual inspection we found that sentences from the English dataset are often code-mixed with Hindi, which makes this dataset challenging and different from all other datasets. Table \ref{tab: cross_validation} indicates that a Word-BiLSTM model struggles to achieve a reasonable PRAUC on this dataset, even when it is trained with labeled instances from the same dataset.

\begin{table*}
\begin{center}
\begin{tabular}{l|cc|ccc|cc}\toprule
Training set & PRAUC & AUC & PR & REC & F1 & TP & FP \\\hline\hline
WA & 0.583 & 0.673 & 0.654 & 0.307 & 0.417 & 1160.7 & 616.7 \\
DBW & 0.566 & 0.648 & \bf{0.664} & 0.166 & 0.265 & 627.3 & 317 \\
Unbiased WA + hate speech from DBW (AR) & 0.605 & 0.674 & 0.533 & \bf{0.684} & \bf{0.598} & \bf{2588.9} & \bf{2283.7} \\
all WA + hate speech from DBW  & \bf{0.645} & \bf{0.716} & 0.659 & 0.49 & 0.562 & 1855.75 & 961.13 \\\bottomrule
\end{tabular}
\caption{Addition of more examples of hate speech is comparable to unbiasing the data set. PRAUC values reported for WA and AR are slightly different from the ones reported in Table \ref{tab: cross_validation}, because we perform in-domain cross validation in that table.}
\label{tab:arango_ablation}
\end{center}
\end{table*}

\section{Experimentation}\label{experiments}
% Description of the work by Arango et al. and why it was important
% We replicate the Arango Work, but their work leaves two questions open:
% a. The metrics make it hard to interpret whether the system has improved.  There is a class imbalance in the data, suggesting that a micro-average of the metrics is a more informative summary.  In many hate speech applications, recall is more important than precision.  For example, in a recommender system, the cost of recommending hate speech is much greater than the cost of making a sub-optimal precision.  Therefore we prefer a system that has higher hate speech recall, at the expense of precision.  
% We evaluate additional metrics (PRAUC, AUC, True positives and false positives, along with both the micro and macro averages) to provide a better comparison.
% b. Is it the sample bias or simply the amount of data that makes the difference?
% We add in the extra data, and indeed, it makes a big difference
%\subsection{Hate Speech Detection Models}
\label{sec:hsd_models}
We consider three different models based on text representation techniques. The first one, \emph{Word-BiLSTM}, is a BiLSTM based model proposed by \citet{agrawal2018deep} and used by \citet{Arango:2019}. The second, \emph{Char-CNN}, is  a Convolutional Neural Network (CNN) that applies convolution over character representations. The third model, \emph{Subword-BERT}, is a fine-tuned BERT \cite{devlin-etal-2019-bert}, which uses subwords to convert text to vectors. For all the models, the validation set was 10\% of the training set (source domain data + weakly labeled data).  
  
%The underlying text representation technique of this model is word embedding. We refer to this model as \emph{Word-BiLSTM}. We also use a Convolutional Neural Network (CNN) based model that applies convolution over character representations. We refer to this model as \emph{Char-CNN}. The third model that we use is a fine-tuned BERT \cite{devlin-etal-2019-bert}, which uses subwords to convert text to vectors. We refer to this model as \emph{Subword-BERT}. 

We show that the domain adaptation approach described above improves results across a variety of models and data sets, even when the text is a mixture of languages and uses character-level substitutions. All the results in this paper are produced by running the same algorithm 10 times with the same hyper-parameters using 10 different random seeds and averaging performance. 

\subsection{Model Details}

The focus of this work is on the domain adaptive data generation, not on the models themselves.  We show in the experimental results section that a different model performs best in each target domain because of the token representation. Character attacks are very common in hate speech and BERT fine-tuning also fails with character level adversarial attacks. We do not propose or advocate any specific model in this paper, as the focus is the data generation, and it is model-agnostic by design.

\label{sec:models}
\textbf{Word-BiLSTM} follows \citet{agrawal2018deep}, who proposed a deep learning model for the detection of cyberbullying, which often involves hate speech. They explored CNN, LSTM, BiLSTM, and BiLSTM with attention architectures with the underlying Glove word embedding representation. The results for all architectures were similar. As we compare our results with \citet{Arango:2019}, we also use the BiLSTM model. The sequence of layers in this architecture is word embedding, then a BiLSTM, then fully connected layers, and finally softmax. The authors used 50-dimensional word vectors and LSTMs (both directions makes it 100 dimensional). We apply Dropout after the BiLSTM and word embedding layers. Even though \cite{Arango:2019} trained the BiLSTM model with the Adam optimizer for 10 epochs, we further create a validation set and follow an early stopping strategy with patience value of 3.

%\paragraph{
\textbf{Char-CNN} is an implementation of the model proposed by \citet{NIPS2015_5782}. This model looks at the input text as a sequence of characters. Given the sequence of character embedding, this model applies six layers of convolution with max-pooling.  Then it applies three fully connected layers with two dropout modules in between them for regularization. The early stopping mechanism was used for this CNN with patience value of 3.% Please refer to \citet{NIPS2015_5782} for a detailed description of the model.      

\textbf{Subword-BERT} uses the BERT\textsubscript{base} model to encode text~\cite{bert2018}. %as shown in Figure \ref{fig:BERT_model}. 
We apply a special token [CLS] at the beginning of the text and another token [SEP] at the end of the text. We take the representation of the [CLS] token from the 12\textsuperscript{th} layer of BERT, which is a 768-dimensional vector and pass it through a Fully Connected (FC) layer. Finally, we apply a softmax activation function on the representation computed by the FC layer to classify. We used a batch size of 32, with a learning rate of 2e-5, and trained the model for three epochs. \citet{bert2018} mentioned that 2-4 epochs of fine-tuning is quite effective for the Glue tasks. We found that training for 3 epochs works best in our setting.

%\vanessa{get the naacl baseline from the slides}
\subsection{Preliminary Experiments}\label{prelim}

The selected datasets provide a platform for creating a challenging domain adaptation setting. We demonstrate this by showing the drop in PRAUC (Area Under the Precision-Recall Curve), when the training and test set are from different datasets compared to when they are from the same dataset, as shown in  Table~\ref{tab: cross_validation}. Note that the diagonal represents testing on a held out set of 10\% of the data, and training on the other 90\%. We used the word-BiLSTM model described in section~\ref{sec:models} for these experiments.

We replicate the results of \citet{Arango:2019}, and further add hate speech examples from DBW without limiting the number of tweets from a single user.  We run the word-BiLSTM model using the hyper-parameters from \citet{Arango:2019}. We report PRAUC and AUC, True Positives and False Positives, alongside precision, recall, and F1 scores reported by \citet{Arango:2019}. The result is shown in Table \ref{tab:arango_ablation}. 
The first two rows show that using WA and DBW alone results in poor performance when adapting to SE. The third and fourth rows show that limiting tweets from users is less effective if we consider PRAUC and AUC, as shown by comparing WA to the unbiased version of WA when adding hate speech examples from DBW to both sets. % add \hdavidson with the biased dataset \waseem -- compared to the unbiased \uwaseem proposed by \citet{Arango:2019}. 

\subsection{Unsupervised Domain Adaptation Setting}
\label{sec:uda_setting}
Unsupervised domain adaptation assumes that only unlabeled data exists in the target domain. To create such a setting, we randomly sample 10\% data from each of the target datasets to create unlabeled data. This resulted in 900, 1095, and 586 randomly selected sentences from SE, GI, and GA datasets, respectively. We do not use the labels of these sentences but use the sentences themselves in the noisy data generation process. The remaining data is used as test set. In the SE, GI, and HA test sets there are 3409, 1097, and 1040 hate speech examples, and 4691, 8752, and 4226 non-hate speech examples, respectively. %There can be several other settings such as taking a larger proportion of data as unlabeled data or creating data folds from test data etc. We do not explore these settings in this paper. 
The data is not truly a uniform random sample from the unlabeled data of the target corpus as it is a part of the original labeled data. However this is a typical limitation of UDA settings.   

To show the effectiveness of our proposed approach, we use AR as the baseline training data, and show the improvements that we obtain by augmenting domain adaptive weakly supervised data with AR. Our technique involves training an Offensive or Target Group (OTG) tagger from AR, and we adapt the sequence tagger implementation of \citet{yang2018ncrf} for this task.

Note that AR consists of unbiased WA and DBW. While the DBW dataset is sampled using a hate speech lexicon taken from \url{hatebase.org}, WA was not sampled in that way. Following our approach described in Section \ref{sec:otg_tagger}, we require to match tokens from a hate speech lexicon to hate speech data for generating training data for the OTG tagger. We only use the DBW portion of the AR dataset for this purpose. We use an n-gram based matching technique to map the tokens from the hate speech lexicon to the 4993 hate speech in DBW. Once we train the OTG tagger with this data, we run the tagger on a large scale \emph{weakly supervised} sentiment analysis dataset provided by \citet{go2009twitter}. This dataset contains 800,000 negative emotion sentences that we convert to hate speech templates using the OTG tagger, as described in Section~\ref{sec:otg_tagger}. 

 Following the approach described in section \ref{sec:approach}, we rank these templates by their similarity to the target domain, select top 10,000 hate and non-hate templates, and convert them to hate and non-hate examples. The value of 10,000 was determined empirically, by tuning it as a parameter on a held out set.

%The value of $k$ was selected based on the performance on the validation set. For a particular value of $k$, we create the training data (strongly labeled sentences + top-$k$ weakly labeled hate and non-hate sentences) and select 10\% of it as validation data. After tuning $k$ we found $k=10000$ results in the best performance across validation sets computed from all the training datasets.  \looseness-1

Experiments described in the previous section indicated that data augmentation from the hate speech class is one of the key factors in reducing bias and adapting to a new domain. Results of the experiments in table \ref{tab:totalEval} show the effectiveness of adding domain adapted, weakly labeled data to the AR data, evaluated on the SE, GI, and HA test sets, respectively. %The evaluation was performed on the three types of models described above. 

%Table \ref{tab:totalEval} shows the performance of character, word and subword based models trained with \uwaseem alone and in combination with our weakly supervised data. 
%The predictions are performed on the test set from the target domain dataset. The test dataset contains 90\% of the original data set, while 10\% of the original dataset was used as unlabeled data to create our weakly supervised dataset. 
%As the weak dataset is domain aligned using unlabeled data from Semeval dataset, we refer to it as \sweak. 

Table \ref{tab:totalEval} shows that the addition of weakly labeled data improves PRAUC, AUC and F1 metrics for all types of models for the hate speech class. The per-class metrics can be inferred from the True Positives and False Positives and the total number of examples in the data set. In particular recall has a larger gain for character and subword models compared to the word-based model.  This is especially notable for the HA data which includes examples that are a code-mix of Hindi and English. %This is a little counter-intuitive as for all the other datasets in Table \ref{tab:gibert18} and Table \ref{tab:hasoc19}, we observe that the weakly labeled dataset boosts recall. 
Another important observation is that although BERT fine-tuning is a strong baseline for text classification tasks, it performs worse than the word embedding BiLSTM model on the SE data. This does not hold for the GI data, where we find that BERT fine-tuning supercedes all the other approaches by a large margin. This could be accounted for by the fact that the GI data is sampled from white-supremacists' forum posts which includes complete grammatical sentences, whereas the SE data is from Twitter. As BERT has been trained on Wikipedia, it models this type of content better.

\begin{table*}
\begin{center}

\begin{tabular}{c|l|l|cc|ccc|cc}\toprule
Target & Model & Training & PRAUC & AUC & PR & REC & F1 & TP & FP \\
Domain & & Data& & & & & & & \\\hline\hline
\parbox[t]{2mm}{\multirow{6}{*}{\rotatebox[origin=c]{90}{\bf{SE}}}} & {Char-CNN} & AR & 0.549 & 0.591 & 0.460 & 0.590 & 0.517 & 2012 & \bf{2358} \\
&  & AR + SE\textsubscript{weak} & \bf{0.558} & \bf{0.646} & \bf{0.496} & \bf{0.748} & \bf{0.597} & \bf{2549} &  2585 \\\cline{2-10}
&Word-BiLSTM  & AR  & 0.605 & 0.674 & 0.533 &\bf{ 0.684} & 0.598 & \bf{2588.9} & 2283.7 \\
& & AR + SE\textsubscript{weak}
&\bf{0.653}  & \bf{0.729} & \bf{0.611}  & 0.652 & \bf{0.631} & 2222 &  \bf{1415} \\\cline{2-10}
&Subword-BERT & AR & 0.599 & 0.675 & \bf{0.551} & 0.637 & 0.591 & 2170 & \bf{1765} \\
& &  AR + SE\textsubscript{weak} & \bf{0.613} & \bf{0.697} & 0.541 & \bf{0.740} & \bf{0.625} &  \bf{2521.5} & 2140 \\\hline\hline
\parbox[t]{2mm}{\multirow{6}{*}{\rotatebox[origin=c]{90}{\bf{GI}}}} & Char-CNN & AR & \bf{0.174} & \bf{0.628} & 0.153 & 0.478 & 0.232 & 524 & 2905 \\
&  & AR + GI\textsubscript{weak} & 0.167 & 0.613 & \bf{0.166} & \bf{0.500} & \bf{0.249} & \bf{548} & \bf{2750} \\\cline{2-10}
& Word-BiLSTM  &  AR & 0.151 & 0.514 & 0.151 & 0.297 & 0.200 & 326 & 1832 \\
&  &  AR + GI\textsubscript{weak} &  \bf{0.225} & \bf{0.660} & \bf{0.213} & \bf{0.442} & \bf{0.288} & \bf{485} & \bf{1787} \\\cline{2-10}
& Subword-BERT & AR & 0.291 & 0.758 & 0.234 & 0.644 & 0.343 & 706 & 2309 \\
&  & AR + GI\textsubscript{weak} & \bf{0.331} & \bf{0.786} & \bf{0.260} & \bf{0.644} & \bf{0.369} & \bf{706.5} & \bf{2019.5} \\\hline\hline
\parbox[t]{2mm}{\multirow{6}{*}{\rotatebox[origin=c]{90}{\bf{HA}}}}& Char-CNN & AR & 0.216 & \bf{0.519} & 0.203 & 0.225 & 0.213 & 234 & \bf{921} \\
&  & AR + HA\textsubscript{weak} & \bf{0.307} & 0.514 & \bf{0.203} & \bf{0.845} & \bf{0.327} & \bf{879} & 3461 \\\cline{2-10}
& Word-BiLSTM & AR & 0.205 & 0.510 & 0.203 & 0.474 & 0.283 & 541.4 & \bf{2130.3} \\
& &  AR + HA\textsubscript{weak} & \bf{0.217} & \bf{0.533} & \bf{0.209} & \bf{0.555} & \bf{0.304} & \bf{577} & 2183 \\\cline{2-10}
& Subword-BERT & AR & \bf{0.209} & 0.525 & \bf{0.218} & 0.254 & 0.234 & 264 & \bf{948} \\
&  &  AR + HA\textsubscript{weak} & 0.208 & \bf{0.526} & 0.205 & \bf{0.851} & \bf{0.331} & \bf{885} & 3434.5 \\\bottomrule
\end{tabular}

\caption{The UDA approach improves over training with source domain dataset, AR, taken from \cite{Arango:2019}. AR is a combination of unbiased WA and hate speech from DBW.  SE\textsubscript{weak}, GI\textsubscript{weak} and HA\textsubscript{weak} indicate the domain-adapted weakly labeled data as described in Section~\ref{sec:approach}. The results are average of 10 runs and the best results are boldfaced.}
%\caption{Domain Adaptation improves hate speech detection results for all models, even with data very different from the source domain as shown in the rows labeled "Hasoc".  Hasoc data is a mix of Hindi and English and is least similar to the training data. $W$ stands for the waseem data, $S$, $G$ and $H$ stand for semeval, gibert and hasoc data sets respectively. The best results are in bold.}
\label{tab:totalEval}
\end{center}
\end{table*}

\begin{table}
\begin{center}
\begin{tabular}{c|c|ccccc}\toprule
\multicolumn{2}{c|}{ } &\multicolumn{5}{c}{Testing Set} \\\cline{3-7}
\multicolumn{2}{c|}{} & \bf{WA} & \bf{DBW} & \bf{SE} & \bf{HA} & \bf{GI} \\\hline\hline
\parbox[t]{2mm}{\multirow{5}{*}{\rotatebox[origin=c]{90}{Training Set}}} & \bf{WA} & \bf{0.768} & 0.199 & 0.561 & 0.198 & 0.103 \\
& \bf{DBW} & 0.390 & \bf{0.465} & 0.525 & 0.191 & 0.079\\
& \bf{SE} & 0.390 & 0.226 & \bf{0.725} & 0.195 & 0.133\\
& \bf{HA} & 0.396 & 0.213 & 0.421 & \bf{0.240} & 0.062\\
& \bf{GI} & 0.384 & 0.275 & 0.455 & 0.172 & \bf{0.404}\\\bottomrule
\end{tabular}
\caption{Cross-dataset performance represented using PRAUC. The same 90/10 train/test split was used in each comparison. In most cases, the results are significantly worse on out-of-domain test data.}\label{tab: cross_validation}
\end{center}
\end{table}

\subsection{Model Adaptation vs. Data Augmentation}
\label{sec:naccl_results}

The model improvements presented in this paper are data-driven, as we increase the model effectiveness by augmenting weakly labeled data with source domain data in the training process. Model-driven approaches, such as ACAN~\cite{acan:2019} take advantage of the unlabeled target domain data in the training process for learning domain-agnostic representations, but they do not use any external data. As ACAN is a strong baseline for UDA for sentiment analysis, we investigate its performance for hate speech detection. ACAN uses Glove word embeddings as the underlying representation, and thus it is comparable to the Word-BiLSTM model used in this paper. Note that the Word-BiLSTM is not trained with any domain alignment objective, but it receives the weakly labeled data as input along with the source domain data.

% \usepackage{multirow}
\begin{table}[!ht]
\resizebox{!}{0.185\linewidth}{
\begin{tabular}{c|l|c|c|c|c|c}
\toprule
\begin{tabular}[c]{@{}c@{}}Target \\ Domain\end{tabular} & Approach & PRAUC          & AUC            & P              & R              & F1             \\
\hline \hline
\multirow{2}{*}{SE}                                      & ACAN     & 0.619          & 0.699          & 0.469          & \textbf{0.936} & 0.625          \\ 
                                                         & Proposed & \textbf{0.653} & \textbf{0.729} & \textbf{0.541} & 0.740          & 0.625          \\ \hline
\multirow{2}{*}{GI}                                      & ACAN     & 0.185          & 0.651          & 0.127          & \textbf{0.933} & 0.224          \\
                                                         & Proposed & \textbf{0.225} & \textbf{0.660} & \textbf{0.213} & 0.442          & \textbf{0.288} \\ \hline
\multirow{2}{*}{HA}                                      & ACAN     & \textbf{0.220} & \textbf{0.548} & 0.206          & \textbf{0.905} & \textbf{0.336} \\
                                                         & Proposed & 0.217          & 0.533          & \textbf{0.209} & 0.555          & 0.304         \\
                                                         \bottomrule
\end{tabular}}
\caption{Comparison of the proposed approach with model-driven domain adaptation approach, ACAN~\cite{acan:2019}}
\label{tab:acan}
\end{table}

Table \ref{tab:acan} shows the performance comparison of our approach and ACAN. For the SE and GI datasets, our proposed approach performs better than ACAN across a variety of evaluation metrics, primarily driven by higher precision. However, ACAN performs better on the HA dataset.  The HA data set is the most dissimilar to the source data, as it includes a code-mixed Hindi and English examples, where Hindi words are transliterated using the English alphabet.  The better performance of model-driven adaptation suggests that model-based approaches may be suitable when the source and target domains are very different. We only use ACAN as a reference point as to the best of our knowledge, there is no work on UDA for hate speech detection.  It is possible that using both in combination would improve the results further. 
% Here are the NaACL Baselines:
% Gibert:
% PR AUC AUC PR REC F1 TP FP
% 0.185 0.651 0.127 0.933 0.224 1023 7030
% Hasoc:
% 0.22 0.548 0.206 0.905 0.336 941.5 3625
% semeval:
% 0.619 0.699 0.469 0.936 0.625 3191.5 3618.5
\section{Discussion and Conclusion}\label{discussion}
The main challenge in hate speech detection is not the bias, but the data imbalance that arises from having a limited set of examples of hate speech because hate speech is generated by few users. 
Even if a large number of examples are sampled from a source such as Twitter, a domain gap exists because of the many linguistic variants, targets of the hate speech, and topics that are vectors of hate. We created a domain-specific hate speech data generator by turning a large collection of weakly supervised negative sentiment sentences into domain adapted hate speech. We demonstrated that the approach improves results over training on data from a different domain, even when bias has been reduced in the original data. 

Although WA was shown to be biased by \citet{Arango:2019}, training with only DBW yields worse performance compared to WA. We didn't experiment with this further by checking if bias exists in the hate speech examples from DBW as well, as it is not our research direction, but Table \ref{tab:data_stat} reflects that DBW has a greater class imbalance compared to WA. Over-sampling the hate speech class in both cases did not resolve the problem.

Training with WA augmented with hate speech examples from DBW results in fewer true positives, compared to training with the unbiased WA data. This suggests that the high precision and low recall is the result of over-fitting to the hate speech of a few users. The overall performance is still close to the unbiased WA dataset, indicating that adding more data from the hate speech class reduces the bias.

The F1 value in the hate speech class reported by~\citet{Arango:2019} trained on the WA data is low compared to our implementation of the same model, indicated in Table \ref{tab:arango_ablation}. We looked at the source code obtained from the authors and found that our implementation differed in three ways: we created a validation set, implemented an early stopping strategy, and did not consider the test data vocabulary while constructing the word embedding table. However, we observed a little change in F1 in the hate speech class when training with unbiased WA. Even though we obtained different results, the gain in terms of F1 with unbiasing is still evident.

A limitation of the data generation approach is that it captures sentences that follow a specific template, requiring two slots for imputing offensive content, rather than just one.  The assumption is that to be hate speech (rather than just offensive content) there must be an offensive descriptor, directed at a subject in the sentence.  In real life, there are myriad ways to express hate, which may not be reflected in this particular template.  The template approach will be most effective when the negative sentiment sentences are topically related to the domain of hate speech.  It will do poorly when the hate speech contains implicit mentions of target groups, or implicit hate.

The template generation process is noisy.  For example, a one-slot negative example (not hate speech) from the actual data is ``I wish i got to ... it with you i miss you and how was the premiere''.  A positive example (hate speech, with two slots) is ``fml so ... for seniority bc of technological ineptness i now have to register for ...''. This does not matter for the purpose of hate speech detection, because the only purpose of the domain-adapted data is to capture topically similar negative sentiment context, which can be made domain-specific with hate tokens.  Further, we select the most topically related context sentences and discard the rest.

%\vanessa{We did not use all combinations of datasets in this paper, as we thought that having one source and three targets is sufficient to show the effectiveness of our approach, and we expect different combinations to yield similar conclusions. We also directly compared our numbers with the recent work of Arango et al. using this setting. We wanted to be comprehensive in terms of the evaluation measures and reported AUC, PRAUC, TP, FP, going beyond the evaluation measures reported in existing hate speech papers. Having all combinations of datasets with all these evaluation measures would have created a clutter of tables in our paper with less opportunity to discuss our results. }

%In all subsequent experiments we use the unbiased waseem% dataset as a baseline dataset and show that augmenting noisy-labeled data with \uwaseem helps to adapt to target domain. We show this across different models and different target datasets.   
 
%Discuss Hasoc - multilingual, domain adaptation especially improves recall with no loss in precision.  Char model better, potentially because of the inclusion of non-western characters in some of the words.

Deep learning is especially suited to hate speech detection because there are very few features that can be crafted that are not dependent on a specific hateful vocabulary, whereas hate speech itself is often considerably more subtle, using no specifically hateful term.  Still, there may be benefit to adding features of the community or social network structure, on the basis that people engaged in hate speech form a community and often coordinate to conduct a campaign of hate.  We also leave to future work combining model-based approaches with data augmentation.

\section{Statement of Ethics}
Although this paper did not require additional human labeling of hate speech, it does use human-labeled data, which was created at an additional cost to the human psyche and is itself harmful to the annotator.  Revealing ways to detect hate speech instructs promoters of hate how to avoid detection.  While identifying individual instances of hate speech may be helpful, it is not sufficient to dismantle coordinated attacks, or communities with a vested interest in promoting and normalizing hate, or to address underlying structural issues that permit the use of hate to harm at-risk communities.

% \section*{Acknowledgments}
% This work was supported in part by the Center for Intelligent Information Retrieval. Any opinions, findings and conclusions
% or recommendations expressed in this material are those of the authors and do not necessarily
% reflect those of the sponsor.
% \bibliographystyle{aaai21}
\bibliography{sample-bibliography}

\end{document}